\documentclass[letterpaper, 10 pt, conference]{ieeeconf}  

\IEEEoverridecommandlockouts                              

\usepackage{graphicx} 
\usepackage{amsmath} 

\usepackage[symbol]{footmisc}

\usepackage{url}
\usepackage{booktabs}
\usepackage{multirow}
\usepackage{hyperref}
\hypersetup{
  colorlinks = true, 
  urlcolor   = magenta, 
  linkcolor  = blue, 
  citecolor  = red 
}

\title{\LARGE \bf
AutoTAMP: Autoregressive Task and Motion Planning with LLMs as Translators and Checkers
}

\author{Yongchao Chen$^{1,2}$, Jacob Arkin$^{1}$, Charles Dawson$^{1}$, Yang Zhang$^{3}$, Nicholas Roy$^{1}$, and Chuchu Fan$^{1}$
\thanks{$^{1}$Massachusetts Institute of Technology. jarkin@mit.edu, cbd@mit.edu, nickroy@csail.mit.edu, chuchu@mit.edu }
\thanks{$^{2}$Harvard University. yongchaochen@fas.harvard.edu}
\thanks{$^{3}$MIT-IBM Watson AI Lab. yang.zhang2@ibm.com}
}

\begin{document}

\maketitle
\thispagestyle{empty}
\pagestyle{empty}

\begin{abstract}
For effective human-robot interaction, robots need to understand, plan, and execute complex, long-horizon tasks described by natural language. Recent advances in large language models (LLMs) have shown promise for translating natural language into robot action sequences for complex tasks. However, existing approaches either translate the natural language directly into robot trajectories or factor the inference process by decomposing language into task sub-goals and relying on a motion planner to execute each sub-goal. When complex environmental and temporal constraints are involved, inference over planning tasks must be performed jointly with motion plans using traditional task-and-motion planning (TAMP) algorithms, making factorization into subgoals untenable. Rather than using LLMs to directly plan task sub-goals, we instead perform few-shot translation from natural language task descriptions to an intermediate task representation that can then be consumed by a TAMP algorithm to jointly solve the task and motion plan. To improve translation, we automatically detect and correct both syntactic and semantic errors via autoregressive re-prompting, resulting in significant improvements in task completion. We show that our approach outperforms several methods using LLMs as planners in complex task domains. See our project website\footnote[4]{https://yongchao98.github.io/MIT-REALM-AutoTAMP/\label{website}} for prompts, videos, and code.
\end{abstract}

\section{INTRODUCTION} \label{sec:introduction}
Providing agents with the ability to find and execute optimal plans for complex tasks is a long-standing goal in robotics. Robots need to not only reason about the task in the environment and find a satisfying sequence of actions but also verify the feasibility of executing those actions given the robot's motion capabilities. This problem is referred to as task and motion planning (TAMP), and there has been considerable research on efficient algorithms \cite{integrated-TAMP}. Classic solutions rely on specifying tasks in a dedicated planning representation, such as PDDL \cite{pddl_reference} or Temporal logics \cite{temporal_logic_reference}, that is both sufficiently expressive to specify task complexities (e.g. constraints on task execution) and amenable to such algorithms \cite{pddl_reference,temporal_logic_reference,manipulation-planning-with-tl,strips-reference}.

While this approach to task specification has been quite successful, directly using these representations requires training and experience, making them poor interfaces for non-expert users. As an alternative, natural language (NL) provides an intuitive and flexible way to describe tasks. Pre-trained large language models (LLMs) have demonstrated surprisingly good performance on many language-related tasks \cite{llms-few-shot-learners}, and there has been an associated burst of research applying them to task execution \cite{tidybot}, task planning \cite{llms-zero-shot-planners, saycan, progprompt, inner-monologue} and TAMP \cite{text2motion, llm-grop}.

Promising early efforts used LLMs as direct task planners \cite{llms-zero-shot-planners} generating a sequence of sub-tasks based on a set of natural language instructions, but these approaches were limited by a lack of feedback and inability to verify whether sub-task sequences are executable. Further research addressed executability by connecting sub-tasks to control policy affordance functions \cite{saycan}, providing environmental feedback of robot actions \cite{inner-monologue}, and interleaving action feasibility checking with LLM action proposals \cite{text2motion}; this last work also addressed long-horizon action dependencies. However, these approaches struggle with complex tasks involving temporally-dependent multi-step actions, action sequence optimization \cite{saycan, inner-monologue}, and constraints on task execution \cite{text2motion}. Furthermore, these frameworks factor the planning problem and use LLMs to infer a task plan separately from the motion plan. In many situations, the task and motion plan must be optimized together to fulfill the task. For instance, when the task is `reach all locations via the shortest path', the order of places to be visited (task planning) depends on the geometry of the environment and the related motion optimization. Unfortunately, we find that LLMs do not seem capable of directly generating trajectories, possibly due to limitations in complex spatial and numerical reasoning \cite{chatgpt_for_robot,llms-still-cannot-plan}.

To benefit from both the user-friendliness of NL and the capabilities of existing TAMP algorithms, we approach the problem by using LLMs to translate from high-level task descriptions to formal task specifications. We are not the first to use LLMs in this way \cite{llm+p,translating-NL-to-PDDL-goals}, but our work addresses some limitations of prior approaches. Previous work translated natural language to Linear Temporal Logics (LTL) \cite{danli}, which only considered the problem of task planning, and PDDL problem descriptions \cite{llm+p} or PDDL goals \cite{translating-NL-to-PDDL-goals}. Here we utilize Signal Temporal Logic (STL) as the intermediary representation, allowing for more expressive constraints than LTL and facilitating integrated task and motion planning as with PDDL \cite{garrett2020pddlstream}.

The LLM translation process can produce malformed (\emph{syntax errors}) and semantically misaligned (\emph{semantic errors}) formal task specifications. To address syntax errors, we adopt an existing iterative re-prompting technique that relies on an external syntax verifier to prompt the LLM with the specific syntactic error for correction \cite{errors-are-useful-prompts}. Unfortunately, the lack of an external verifier makes this technique inapplicable for a semantic misalignment between the original natural language instruction and the translated specification. To address this problem, we contribute a novel autoregressive re-prompting technique that uses an LLM to evaluate whether the generated plan is semantically consistent with the original instruction. We re-prompt the model to check the alignment between the original instruction and the generated plan by providing the context of the instruction, the generated STL, and the output of the planner. We conduct comprehensive experiments in challenging 2D task domains, including several multi-agent tasks, and find that our approach outperforms direct LLM planning for tasks with hard geometric and temporal constraints. We show that, when combined with automatic syntactic correction, our technique significantly improves task success rates. We conduct an ablation study over the translation step by integrating a fine-tuned NL-to-STL model \cite{NL2TL} with the AutoTAMP framework and show that GPT-4 few-shot learning is competitive with fine-tuning. In addition to our code, we publish a dataset of 1400 test cases consisting of the language instructions, environments, generated STL, and planner trajectory outputs. We conclude that in-context learning with pre-trained LLMs is well suited for language-to-task-specification translation for solving TAMP problems.

\section{PROBLEM DESCRIPTION} \label{sec:problem-description}
As shown in Figure \ref{fig:three-methods}, we aim to convert a natural language instruction, including spatial and temporal constraints, into a motion plan for a robot encoded as a set of timed waypoints, e.g., $(x_i, y_i, t_i)$. The environment state is encoded as set of named obstacles described as polygons and is provided as additional context. Our task is to generate a constraint-satisfying trajectory based on the given instruction and the environment state. The robot must not surpass its maximum velocity, and the total operation time should not exceed the task time limit. We assume that the full trajectory is a linear interpolation between the timed waypoints; complex trajectories can be specified by dense waypoint sequences. 

\section{METHODS} \label{sec:methods}
\begin{figure}[t]
  \vspace{1mm}
  \centering
  \includegraphics[width=1.0\linewidth]{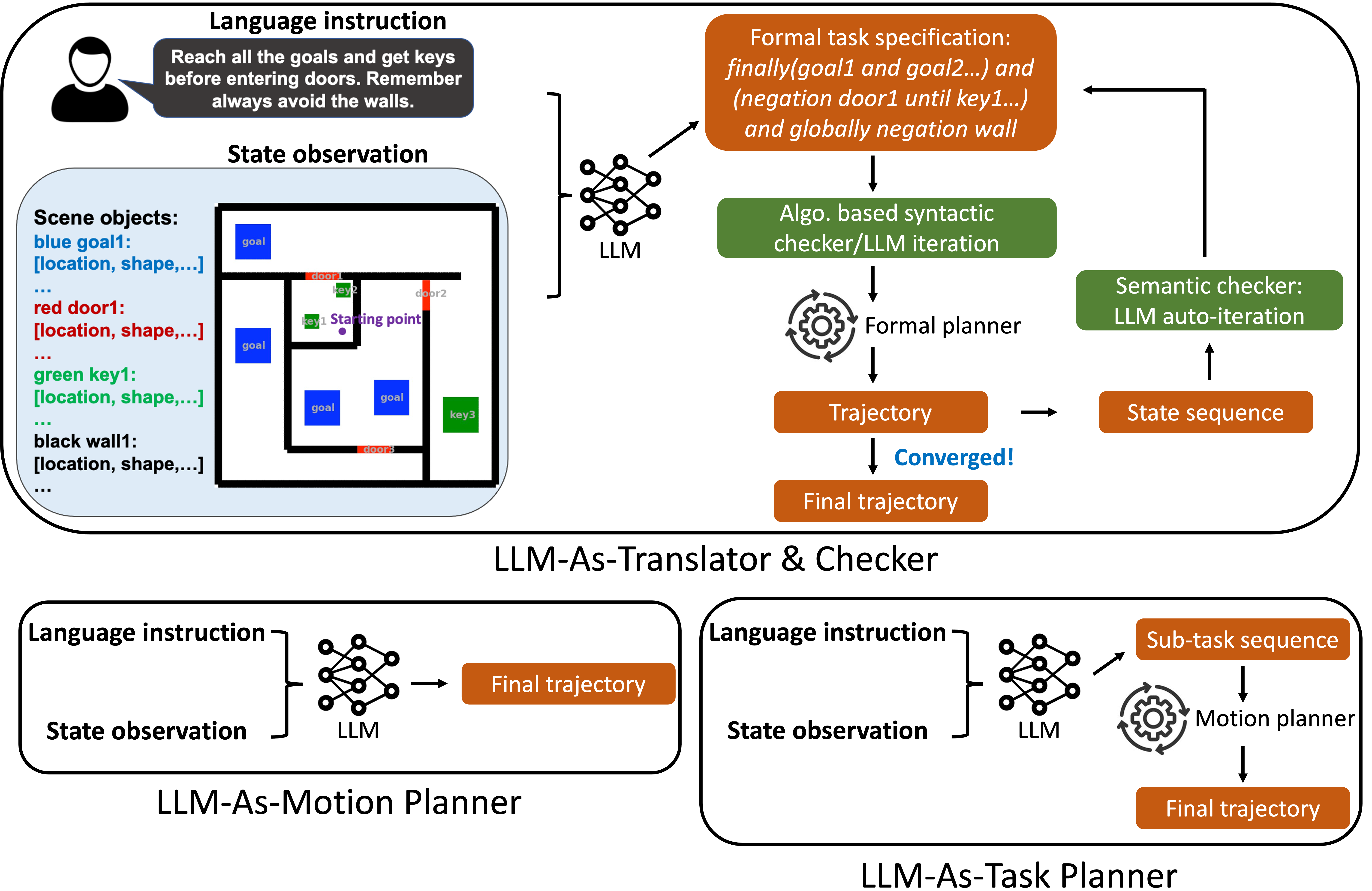}
  \caption{Illustration of different approaches applying LLMs for task and motion planning; our work contributes the LLM-As-Translator \& Checker approach. Each approach accepts a natural language instruction and environment state as input and outputs a robot trajectory.}
  \label{fig:three-methods}
\end{figure}

\begin{figure*}[t]
  \centering
  \includegraphics[width=0.7\linewidth]{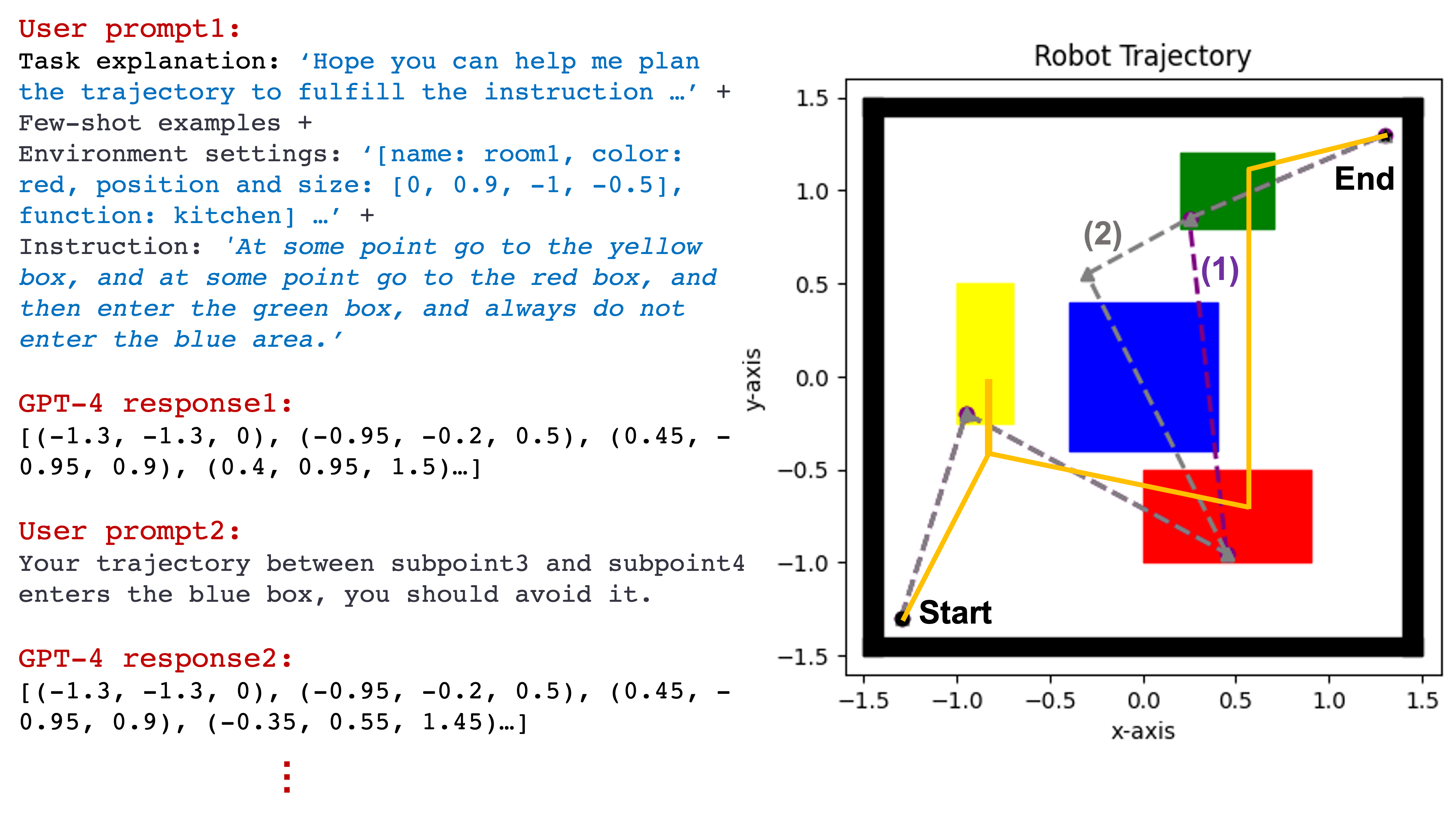}
  \caption{GPT-4 failure case for direct end-to-end trajectory planning. The orange line shows the correct path obeying the instruction. The purple and gray dashed lines show the trajectories from GPT-4 after first and second prompts, respectively. GPT-4 generates a list of $(x, y)$ locations with associated timestamps. The initial prompt  describes the language modeling task, environment state, and instruction. Each object is a rectangle described by $(x, y)$ boundaries.}
  \label{fig:gpt-failure}
\end{figure*}

Figure~\ref{fig:three-methods} illustrates three of the approaches we compare in our work, each using LLMs in some capacity. Each takes as input (1) a text-based representation of the global environment state, (2) in-context examples for few-shot learning, and (3) a natural language instruction. The LLM-As-Translator \& Checker approach is the contribution of this paper. Details and examples of context for prompting and re-prompting can be found in our code repository\footref{website}.

\subsection{LLM End-to-end Motion Planning} \label{subsec:methods-e2e-motion-planning}
One natural idea is to use an LLM for both task and motion planning by directly generating a trajectory for a given language instruction; we refer to this as LLM End-to-end Motion Planning. In cases where the generated trajectory violates constraints, we re-prompt the model with the constraint violation to produce another trajectory, allowing up to five such re-prompts. Figure~\ref{fig:gpt-failure} shows this pipeline, including a specific failure case with two constraint-violating trajectories. The LLM End-to-end Motion Planning violates the constraints multiple times even with direct correction re-prompts. This results from the poor spatial and numerical reasoning abilities of LLMs.

\subsection{LLM Task Planning} \label{subsec:methods-llm-task-planning}
A more common approach is to use an LLM to handle the task planning by directly generating a sequence of sub-tasks from a given language instruction; we refer to this as LLM Task Planning. To generate a final trajectory, the sub-tasks are handled by an independent motion planner. In this work, these sub-tasks are limited to navigation actions, and the motion planning is handled by the STL planner used by our proposed approach; this permits fair comparison of results across methods. Each sub-task is converted to STL to be consumed by the planner. We evaluate and compare against three methods that each use LLMs for task planning: (1) Naive Task Planning, (2) SayCan, and (3) LLM Task Planning + Feedback.

\textbf{Naive Task Planning}\quad As proposed by \cite{llms-zero-shot-planners}, we evaluate using LLMs to generate the entire sub-task sequence without checking for executability.

\textbf{SayCan}\quad Alternatively, an LLM can be iteratively prompted to generate each subsequent sub-task conditioned on the previous sub-tasks in the sequence. The next sub-task can be selected from the top K candidates by combining the language model likelihood with a feasibility likelihood of the candidate action and choosing the most-likely next sub-task, as proposed by \cite{saycan}. We set K to 5 in our evaluations.

\textbf{LLM Task Planning + Feedback}\quad A third task planning method combines full sequence generation with feasibility checking to both find sub-task sequences that satisfy the full task and verify their feasibility before execution. For any infeasible sub-tasks, the LLM can be re-prompted with feedback about the infeasible actions to generate a new sub-task sequence. This is similar to the hierarchical method proposed by \cite{text2motion} but with feedback for re-prompting.

\subsection{Autoregressive LLM Specification Translation\&Checking + Formal Planner} \label{subsec:methods-autotamp}
\begin{figure}[h]
  \vspace{1mm}
  \centering
  \includegraphics[width=0.8\linewidth]{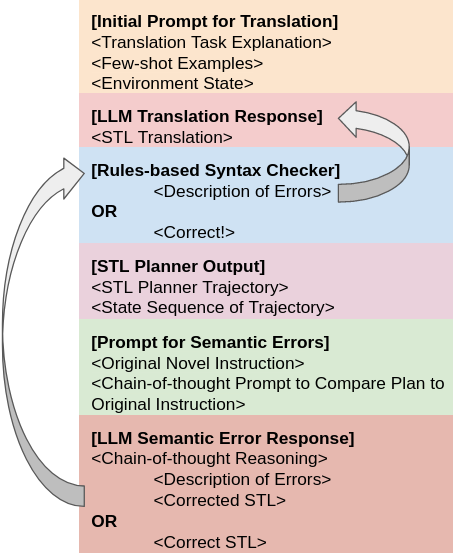}
  \caption{High-level structure of the prompt used for AutoTAMP. The arrow on the right indicates re-prompting for syntax error correction. The arrow on the left indicates re-prompting in cases of semantic errors.}
  \label{fig:prompt-outline}
\end{figure}

Unlike LLM Task Planning, our approach translates NL to STL with an LLM and then plans the trajectory with an STL planner, as shown in Figure~\ref{fig:three-methods}. We include two re-prompting techniques to improve translation performance: one for syntactic errors and another for semantic errors. By ``semantic error", we mean a misalignment between the intended task described in natural language and the STL expression to which it is translated. Figure \ref{fig:prompt-outline} shows the structure of the context for re-prompting the model for semantic error correction; we include a full prompt example in our code repository\footref{website}.

\textbf{Signal Temporal Logic Syntax}\quad In this work, we use STL \cite{stl-reference} as a formal task specification that supports continuous real-time constraints suitable for time-critical missions. An STL formula is defined recursively according to the following syntax:
\begin{multline}
\phi ::= \pi^\mu\;|\;\neg\phi\;|\;\phi\land\varphi\;|\;\phi\lor\varphi\;|\;\textbf{F}_{[a,b]}\phi\;|\;\textbf{G}_{[a,b]}\phi\;
|\;\phi\textbf{U}_{[a,b]}\varphi
\end{multline}
where $\phi$ and $\varphi$ are STL formulas, and $\pi^\mu$ is an atomic predicate. $\neg$ (negation), $\land$ (and), $\lor$ (or), $\Rightarrow$ (imply), and $\Leftrightarrow$ (equal)) are logical operators. $\textbf{F}_{[a,b]}$ (eventually/finally), $\textbf{G}_{[a,b]}$ (always/globally), and $\textbf{U}_{[a,b]}$ (until) are temporal operators with real-time constraints $t \in [a, b]$. The action primitives in this work are 'enter(room\_name)' and 'not\_enter(room\_name)'.

\textbf{STL Trajectory Planner}\quad We use a state-of-the-art multi-agent STL planner \cite{sun2022MILP} that uses piece-wise linear reference paths defined by timed waypoints to recursively encode the constraints expressed in the provided STL expression. It defines the validity of an STL formula with respect to a trajectory and then optimizes the trajectory to maximize the validity. The planner not only searches for a sub-task sequence but also optimizes the time efficiency under dynamical constraints of robot maximum velocity. Here we assume that the locations and shapes of all the objects/rooms in the whole environment are known, which serves as the environment information to the STL planner.

\textbf{Syntactic Checking \& Semantic Checking}\quad Open-loop translation can suffer from syntactic and semantic errors. We use two re-prompting techniques to automatically correct such errors. Like \cite{errors-are-useful-prompts}, we use a verifier to check for syntax errors (we use a simple rules-based STL syntax checker); any errors are provided as feedback when re-prompting the LLM to generate corrected STL. We repeat until no errors are found (up to five iterations). For semantic errors, we propose a novel autoregressive re-prompting technique; we provide the STL planner's generated state sequence (i.e., [[in(road), 0], [in(red kitchen), 0.5], [in(blue restroom2), 1.2],...]) as context alongside the original instruction and ask the LLM to check whether the plan aligns with the instruction's semantics. If it does not, the LLM is prompted to modify the STL, which repeats the syntactic and semantic re-prompting. This process terminates in the case of no detected errors or no change in STL (up to three iterations). The structure of the semantic error prompt is shown in Figure \ref{fig:prompt-outline}; full example prompts can be found in our code repository\footref{website}.

\section{EXPERIMENTAL DESIGN} \label{sec:experimental-design}
Each task scenario is set in a 2D environment and entails navigation of one or more robots; the robots have extent in the environment and are initialized with varying start positions. Each environment consists of regions with shapes, locations, and properties (e.g., color, name, function). For each method, the LLM is initially prompted with a description of the language task (e.g. task planning or translation) and five in-context examples for that task. To mitigate variance across prompts, we initially tested six different sets of examples for each method and chose the one that performed best. Through this testing, we found that the variance over prompts was insignificant relative to overall performance.

We evaluated the different methods described in Section \ref{sec:methods} across six different task scenarios (three single-agent and three multi-agent) with different combinations of geometric and temporal constraints. For each scenario description below, we indicate the presence of these constraints below with G and T, respectively. For each method, we evaluate performance with both GPT-3 and GPT-4 as the LLM. Note that in multi-agent scenarios, we do not test SayCan or LLM Task Planning + Feedback because these methods are not straight-forwardly adaptable for multiple agents. For multi-agent tasks, the agents are assigned a subtask and a time for completion at each time step; since the time for completion is often different, it is not obvious how/when to check and provide feedback. We also terminate and report failure for test cases that take more than 90 minutes. We automatically check resulting trajectories via hard-coded checkers. The full set of experiments took two weeks using four 16-core CPUs; the cost of LLM API calls for evaluating all of the approaches was $\sim$1500 USD.

\begin{figure}[t]
  \centering
  \includegraphics[width=1.0\linewidth]{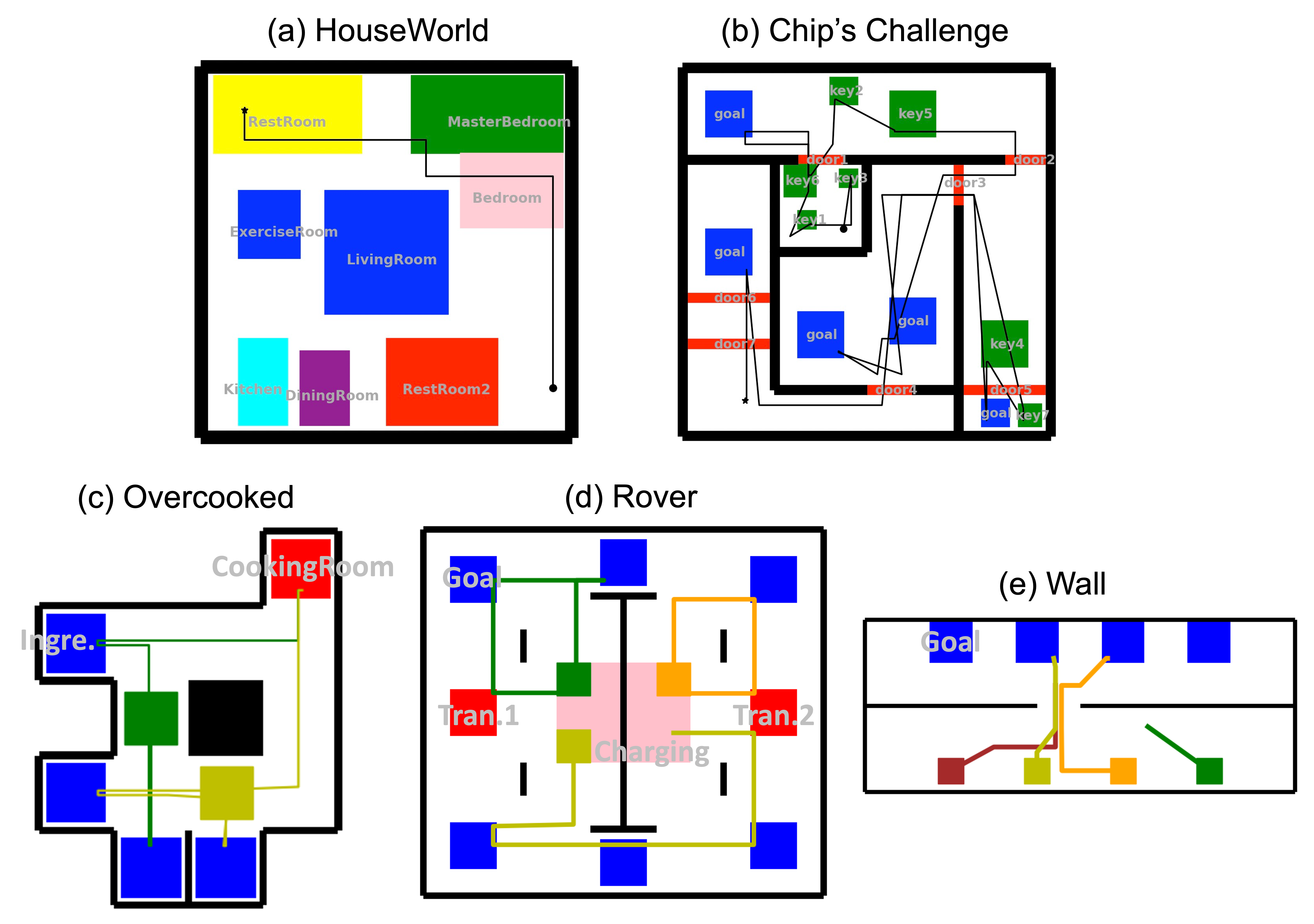}
  \caption{HouseWorld and Chip's Challenge are single-agent scenarios. Overcooked, Rover, and Wall are multi-agent scenarios. The black square in Overcooked is inadmissible. The lines indicate the correct trajectories following the instructions. For the HouseWorld and Chip's Challenge environments, the black round dot and pentagonal dot indicate the start and end positions, respectively.}
  \label{fig:experimental-scenarios}
\end{figure}

\textbf{HouseWorld1 (single-agent)}\quad As shown in Figure~\ref{fig:experimental-scenarios}(a), this is a house environment from \cite{ltlmop}. We first manually constructed 10 different instructions of varying complexity before prompting GPT-4 to paraphrase each into 9 differently worded instructions with the same meaning, resulting in 100 total instructions for this environment. For each instruction, we randomly initialize between two start-end  position pairs for 200 total test cases. For this scenario, we do not impose a hard time constraint for the planned trajectory.

\textbf{HouseWorld2 (T, single-agent)}\quad This scenario is identical to HouseWorld1, but each planned trajectory is subjected to a hard time constraint. This time limit is pre-determined by completing the correct trajectory with 0.8 maximum velocity.

The remaining task scenarios were designed with specific rules and goals for the agent(s) to follow. For each scenario, GPT-4 was used to paraphrase the original description into 20 unique variants with the same meaning, which are further checked by humans. We instantiate three different instances of the environment for each scenario and randomize five different start/end location pairs for a total of 300 test cases.

\textbf{Chip's Challenge (G, single-agent)}\quad Figure~\ref{fig:experimental-scenarios}(b) shows a scenario inspired by Chip's Challenge, a classic puzzle solving game with strict geometric and logical constraints. The robot must reach all goal regions (blue) but must acquire a unique key to pass through the corresponding door.

\textbf{Overcooked (G \& T, multi-agent)}\quad Figure~\ref{fig:experimental-scenarios}(c) shows a scenario inspired by Overcooked, a popular cooking simulation game with strict time constraints. The agents must cooperatively gather ingredients and return to CookingRoom in a limited time. The multi-agent motion planning is challenged by limited space for agents to maneuver.

\textbf{Rover (G \& T, multi-agent)}\quad Figure~\ref{fig:experimental-scenarios}(d) is a scenario used by \cite{sun2022MILP}. Multiple agents must reach each observation region (blue) before transmitting their observations from a red region, all while subjected to time and energy constraints.

\textbf{Wall (G \& T, multi-agent)}\quad Figure~\ref{fig:experimental-scenarios}(e) is also from \cite{sun2022MILP}. Multiple agents must occupy each goal region (blue) while subject to a time constraint and a maneuver bottleneck.

\section{RESULTS} \label{sec:results}
\begin{table*}[]
\vspace{2mm}
\caption{Task success rates for \textbf{single-agent} scenarios. Each scenario's constraints are listed in the table.}
\resizebox{\textwidth}{!}{
\begin{tabular}{|lll|ccc|}
\hline
    &   &   & \multicolumn{1}{c|}{\textbf{HouseWorld1}} & \multicolumn{1}{c|}{\textbf{HouseWorld2}} & \textbf{Chip's Challenge} \\ \cline{4-6} 
    &   &   & \multicolumn{1}{l|}{Soft Time Constraint} & \multicolumn{1}{l|}{Hard Time Constraint} & \multicolumn{1}{l|}{Hard Geometric Constraints} \\ \hline
\multicolumn{1}{|l|}{\multirow{7}{*}{\rotatebox[origin=c]{90}{GPT-3}}}
    & \multicolumn{1}{l|}{\textbf{LLMs as Motion Planners}} & End-to-end Motion Planning
        & 0.0\%   & 0.0\%   & 0.0\%   \\ \cline{2-6} 
    \multicolumn{1}{|l|}{}  & \multicolumn{1}{l|}{\multirow{3}{*}{\textbf{LLMs as Task Planners}}} & Task Planning (naive)
        & 74.0\% & 36.0\% & 0.0\% \\
    \multicolumn{1}{|l|}{} & \multicolumn{1}{l|}{} & SayCan
        & 75.5\% & 36.0\% & 0.0\% \\
    \multicolumn{1}{|l|}{} & \multicolumn{1}{l|}{} & Task Planning (feedback)
        & \textbf{79.0\%} & 40.0\% & 0.0\% \\ \cline{2-6} 
    \multicolumn{1}{|l|}{} & \multicolumn{1}{l|}{\multirow{3}{*}{\textbf{LLMs as Translators}}} & No Corrections
        & 28.0\% & 27.0\% & 29.0\% \\
    \multicolumn{1}{|l|}{} & \multicolumn{1}{l|}{} & Syntax
        & 49.0\% & 47.0\% & 66.0\% \\
    \multicolumn{1}{|l|}{} & \multicolumn{1}{l|}{} & Syntax + Semantics (AutoTAMP)
        & 62.0\% & \textbf{62.0\%} & \textbf{74.3\%} \\ \toprule
\multicolumn{1}{|l|}{\multirow{7}{*}{\rotatebox[origin=c]{90}{GPT-4}}}
    & \multicolumn{1}{l|}{\textbf{LLMs as Motion Planners}} & End-to-end Motion Planning
        & 9.5\% & 9.5\% & 0.0\% \\ \cline{2-6} 
    \multicolumn{1}{|l|}{} & \multicolumn{1}{l|}{\multirow{3}{*}{\textbf{LLMs as Task Planners}}} & Task Planning (naive)
        & 90.0\% & 45.0\% & 0.0\% \\
    \multicolumn{1}{|l|}{} & \multicolumn{1}{l|}{} & Saycan
        & 90.0\% & 47.5\% & 0.0\% \\
    \multicolumn{1}{|l|}{} & \multicolumn{1}{l|}{} & Task Planning (feedback)
        & \textbf{92.0\%} & 49.0\% & 0.0\% \\ \cline{2-6} 
    \multicolumn{1}{|l|}{} & \multicolumn{1}{l|}{\multirow{3}{*}{\textbf{LLMs as Translators}}} & No Corrections
        & 43.5\% & 42.0\% & 42.7\% \\
    \multicolumn{1}{|l|}{} & \multicolumn{1}{l|}{} & Syntax
        & 59.5\% & 59.0\% & 70.0\% \\
    \multicolumn{1}{|l|}{} & \multicolumn{1}{l|}{} & Syntax + Semantics (AutoTAMP)
        & 82.5\% & 82.0\% & \textbf{87.7\%} \\
    \multicolumn{1}{|l|}{} & \multicolumn{1}{l|}{} & NL2TL + Syntax + Semantics
        & - & \textbf{83.5\%} & 86.0\% \\ \hline
\end{tabular}
}
\label{tab:results-table-single-agent}
\end{table*}

\begin{table*}[]
\caption{Task success rates for \textbf{multi-agent} scenarios. Each situation has hard constraints on time and geometry.}
\centering
\begin{tabular}{|lll|ccc|}
\hline
    &   &   & \multicolumn{1}{c|}{\textbf{Overcooked}} & \multicolumn{1}{c|}{\textbf{Rover}} & \textbf{Wall} \\ \cline{4-6} 
    &   &   & \multicolumn{3}{l|}{Hard Time \& Geometric Constraints} \\ \hline
\multicolumn{1}{|l|}{\multirow{5}{*}{\rotatebox[origin=c]{90}{GPT-3}}}
    & \multicolumn{1}{l|}{\textbf{LLMs as Motion Planners}} & End-to-end Motion Planning
        & 0.0\% & 0.0\% & 0.0\% \\ \cline{2-6} 
    \multicolumn{1}{|l|}{} & \multicolumn{1}{l|}{\textbf{LLMs as Task Planners}} & Task Planning (naive)
        & 13.3\% & 0.0\% & 7.0\% \\ \cline{2-6} 
    \multicolumn{1}{|l|}{} & \multicolumn{1}{l|}{\multirow{3}{*}{\textbf{LLMs as Translators}}} & No Corrections
        & 25.0\% & 22.0\% & 74.0\% \\
    \multicolumn{1}{|l|}{} & \multicolumn{1}{l|}{} & Syntax Corrections
        & 70.0\% & 35.0\% & 85.0\% \\
    \multicolumn{1}{|l|}{} & \multicolumn{1}{l|}{} & Syntax + Semantic Corrections (AutoTAMP)
        & \textbf{89.0\%} & \textbf{60.7\%} & \textbf{89.7\%} \\ \toprule
\multicolumn{1}{|l|}{\multirow{5}{*}{\rotatebox[origin=c]{90}{GPT-4}}}
    & \multicolumn{1}{l|}{\textbf{LLMs as Motion Planners}} & End-to-end Motion Planning
        & 5.0\% & 0.0\% & 6.0\% \\ \cline{2-6} 
    \multicolumn{1}{|l|}{} & \multicolumn{1}{l|}{\textbf{LLMs as Task Planners}} & Task Planning (naive)
        & 17.0\% & 0.0\% & 47.0\% \\ \cline{2-6} 
    \multicolumn{1}{|l|}{} & \multicolumn{1}{l|}{\multirow{3}{*}{\textbf{LLMs as Translators}}} & No Corrections
        & 85.0\% & 46.0\% & 95.0\% \\
    \multicolumn{1}{|l|}{} & \multicolumn{1}{l|}{} & Syntax Corrections
        & 94.0\% & 67.0\% & 95.0\% \\
    \multicolumn{1}{|l|}{} & \multicolumn{1}{l|}{} & Syntax + Semantic Corrections (AutoTAMP)
        & \textbf{100.0\%} & 79.0\% & \textbf{100.0\%} \\
    \multicolumn{1}{|l|}{} & \multicolumn{1}{l|}{} & NL2TL + Syntax + Semantic Corrections
        & \textbf{100.0\%} & \textbf{79.7\%} & \textbf{100.0\%} \\ \hline
\end{tabular}
\label{tab:results-table-multi-agent}
\end{table*}

We report the task success rates for the single-agent and multi-agent scenarios in Table~\ref{tab:results-table-single-agent} and Table~\ref{tab:results-table-multi-agent}, respectively. For HouseWorld1 (Figure~\ref{fig:experimental-scenarios}(a)) with no hard time constraint, we find that all methods using LLMs as task planners outperform our approach; whereas our approach can fail due to translation errors, this environment permits direct trajectories between any two positions and thus lacks geometric challenges that direct task planning methods will struggle with. When adding a strict time constraint (HouseWorld2), we see that such methods perform much worse while AutoTAMP's success rate persists. For the other tasks that include geometric constraints, LLM End-to-end Motion Planning and Naive Task Planning both perform quite poorly. Unsurprisingly, we observe a general trend that GPT-4 outperforms GPT-3.

We find that most failures for LLM Task Planning methods result from task execution time violation and sequencing of actions for long-horizon tasks. For example, Chip's Challenge requires the robot to efficiently collect keys for future doors. Also, the Naive Task Planning method fails to avoid collisions in the multi-agent scenarios. Failures for methods that translate to STL primarily are due to incorrect translation; while our re-prompting techniques help address this issue, there remain cases of poor translation.

\textbf{Ablation Studies}\quad In Table~\ref{tab:results-table-single-agent} and Table~\ref{tab:results-table-multi-agent}, we evaluate the impact of syntactic and semantic error correction on using LLMs to translate to STL. The results show that translation with no error correction has modest success across task scenarios, but both syntactic and semantic error correction significantly improve performance; this trend is present across all scenarios. We also evaluate replacing a pre-trained LLM for translation with a state-of-the-art modular translation pipeline, NL2TL, that uses a smaller LLM (T5-large) fine-tuned on a multi-domain corpus of 30K examples of instructions paired with their corresponding temporal logic expressions \cite{NL2TL}; the error correction steps were still performed by GPT-4. Integrating NL2TL performs similarly to using a pre-trained LLM for translation, providing a modest improvement in HouseWorld2 and Rover. We note that incorporating the two re-prompting techniques for error correction is competitive with fine-tuning since we do not rely on additional data or training.

\textbf{3D Simulation} In supplemental videos, we demonstrate plans generated via AutoTAMP in two 3D simulated environments: a drone navigation scenario that requires reasoning about height, and a tabletop color sorting manipulation scenario. We did not incorporate the semantic check for these demos. The STL planner is directly applicable to the drone scenario using timed waypoints, as done in the 2D experiments. For manipulation tasks, we integrated a simple discrete planner to handle the dynamics mode transitions. We discuss this more in Section \ref{sec:conclusion}.

\textbf{Physical Demonstrations} We demonstrate AutoTAMP on physical differential-drive robots via the remotely-accessible Robotarium platform \cite{robotarium} for the Overcooked, Rover, Wall, and Chip's Challenge scenarios. We track the planned trajectories using a low-level controller that also includes a control barrier function to prevent collisions between robots. This controller and the underlying nonlinear dynamics induce a tracking error; we account for this by padding obstacles at planning time. Obstacles are displayed in the robot workspace using an overhead projector. These physical demos provide evidence that our method can be applied to real-world navigation task and motion planning. They are included as part of supplemental videos.

\section{RELATED WORK} \label{sec:related-work}
\textbf{Task and Motion Planning} Planning for robotics involves both high-level, discrete planning of tasks \cite{strips-reference} and low-level continuous planning of motions \cite{lavalle2006planning}; solving these simultaneously is referred to as task and motion planning \cite{integrated-TAMP}. Modern approaches either attempt to satisfy the motion constraints prior to action sequencing \cite{ferrer2017combined,garrett2018ffrob,akbari2016task}, find action sequences then satisfy the motion constraints \cite{lagriffoul2016combining,wolfe2010combined,srivastava2014combined,garrett2020pddlstream}, or interleave these steps \cite{colledanchise2019towards,kaelbling2013integrated,fernandez2018scottyactivity}. For tasks specified in temporal logic, existing methods either use multi-layer planning \cite{he2015towards}, like the aforementioned approaches, or direct optimization via a mixed-integer linear program \cite{katayama2020fast,sun2022MILP} or a non-linear program \cite{takano2021continuous}. Our work focuses on translating natural language to STL, relying on \cite{sun2022MILP} as a TAMP solver, but can be integrated with other STL-based planners.

\textbf{LLMs for TAMP} Recent claims about the impressive reasoning capabilities of LLMs \cite{llms-few-shot-learners,llms-zero-shot-reasoners} have led to interest in such models for task and motion planning. One approach is to directly use LLMs as planners \cite{llms-zero-shot-planners, saycan, text2motion, inner-monologue, tidybot, progprompt, llm-grop}. Initial work showed that zero-shot generation of an action sequence from a high-level task description had relatively poor executability, but few-shot in-context learning, constraining output to admissible actions, and iterative action generation significantly improved performance \cite{llms-zero-shot-planners}. Subsequent efforts grounded the primitive actions to motion control policies, using affordance functions to guide LLM-based task planning \cite{saycan} and TAMP \cite{text2motion}, also adding feedback \cite{inner-monologue}. Other work focused on how prompting can inform task execution\cite{tidybot, llm-grop}. Despite these successes, however, there is evidence that LLMs perform poorly on more realistic tasks \cite{llms-still-cannot-plan, pddl-planning-with-llms}, motivating different approaches. While we are interested in LLMs for TAMP, our work does not directly use LLMs as planners.

\textbf{Translating Language to Task Representations} A natural alternative is to rely on dedicated planners by mapping from natural language to a planning representation. There is a rich history of parsing natural language into formal semantic representations \cite{zettlemoyer2005learning, zettlemoyer2007online, wong2006learning}, of which we only provide a relatively small sampling. The robotics community adopted parsing and other techniques to map language to such representations as lambda calculus \cite{what-to-do-and-how-to-do-it, artzi2013weakly}, motion planning constraints \cite{natural-language-planner-interface}, linear temporal logic \cite{verifiable-grounding-of-instructions,seq2seq-grounding,patel2020grounding,translating-structured-english}, and signal temporal logic \cite{deep-stl,interactive-stl}, among others \cite{bonial2019abstract}. We refer readers to \cite{tellex2020robots} for a more thorough review. To address challenges of data availability, task generalization, linguistic complexity, common sense reasoning, and more, recent work has applied LLMs to this translation problem. Modular approaches have used LLMs to extract referring expressions with corresponding logic propositions to then construct a full temporal logic specification \cite{lang2ltl,NL2TL}. Relying on LLMs for direct translation, other work has mapped from language to PDDL goals \cite{translating-NL-to-PDDL-goals} or full PDDL problems \cite{llms-construct-and-utilize-world-models-for-task-planning, llm+p}. Our work similarly translates to a task specification, but we can represent complex constraints (e.g. temporal), and we introduce a novel mechanism for automatic detection and correction of semantic errors. An interesting alternative maps language to code \cite{code-as-policies}, which is highly expressive but does not easily optimize or provide behavior guarantees for long-horizon tasks.

\textbf{Re-prompting of LLMs} The quality of LLM output is greatly improved with useful context, such as few-shot in-context learning for novel tasks \cite{llms-few-shot-learners}. LLMs for TAMP are typically also provided task-relevant information, such as environment state or admissible actions \cite{progprompt}. Re-prompting with additional context based on LLM output has been shown to be extremely beneficial, such as with iterative action generation \cite{llms-zero-shot-planners}, environmental feedback \cite{inner-monologue}, inadmissible actions \cite{llms-zero-shot-planners, saycan, text2motion}, unmet action preconditions \cite{corrective-reprompting, llms-construct-and-utilize-world-models-for-task-planning}, code execution errors \cite{generalized-planning-in-pddl-domains}, and syntactic errors in structured output \cite{errors-are-useful-prompts}. Our work uses the same syntactic correction re-prompting technique as \cite{errors-are-useful-prompts}, but we also introduce automatic detection and correction of semantic errors via re-prompting.

\section{CONCLUSION} \label{sec:conclusion}
This paper presented AutoTAMP, a framework for using pre-trained LLMs as both (1) translators from language task descriptions to formal task specifications (e.g. STL) via few-shot in-context learning and (2) checkers of syntactic and semantic errors via corrective re-prompting, in which we contributed a novel autoregressive re-prompting technique for semantic errors. Our experimental results show using LLMs to translate to task specifications that can be solved via a formal planner outperforms approaches that use LLMs directly as planners when handling tasks with complex geometric and temporal constraints.

We note a few limitations. First, though our results rely on using the best prompt out of several candidates, alternatives may elicit better performance. However, we expect the trends between methods to persist even with better prompts, supporting the conclusion that LLMs are not well suited for directly solving complex TAMP. Second, the cost of planning time is high, especially when there are multiple iterations of re-prompting. Further work is needed to address the runtime of formal planners and LLM inference. Third, the STL planner used in this work is not immediately applicable to manipulation tasks due to the optimization methods used in the planner; however, our approach does not depend on this specific planner, and we believe it can be integrated with STL planners more suitable for such TAMP domains. 

\section{ACKNOWLEDGEMENTS} \label{sec:ACKNOWLEDGEMENTS}
This work was supported by ONR under Award N00014-22-1-2478, Army Research Laboratory under DCIST CRA W911NF-17-2-0181, and MIT-IBM Watson AI Lab. This article solely reflects the conclusions of its authors.

\clearpage






\bibliographystyle{IEEEtran.bst}
\bibliography{references}

\end{document}